\documentclass[journal]{IEEEtran}

\usepackage{ifpdf}
\usepackage{amsmath,amsfonts}
\usepackage{amssymb}  
\usepackage{algorithmic}
\usepackage{algorithm}
\usepackage{array}
\usepackage[caption=false,font=footnotesize]{subfig}
\usepackage{textcomp}
\usepackage{stfloats}
\usepackage{url}
\usepackage{verbatim}
\usepackage{graphicx}
\usepackage{cite}
\usepackage{booktabs}
\usepackage[export]{adjustbox}
\usepackage{bm}
\usepackage{nccmath}
\usepackage{mathtools}
\usepackage{hyperref}
\usepackage{tensor}
\usepackage[dvipsnames]{xcolor}
\usepackage{units}
\usepackage{tikz-cd}
\usepackage{bbm}
\usepackage{tikz}

\newcommand{\R}{\mathbb{R}}

\begin{document}

\title{Hands-free Telelocomotion of a Wheeled Humanoid toward Dynamic Mobile Manipulation via Teleoperation}


\author{{Amartya Purushottam$^1$, Yeongtae Jung$^2$, Kevin Murphy$^2$, Donghoon Baek$^2$, Joao Ramos$^{1,2}$}
\thanks{The authors are with the $^1$Department of Electrical and Computer Engineering and the $^2$Department of Mechanical Science and Engineering at the University of Illinois at Urbana-Champaign, USA.}
\thanks{This work is supported by the National Science Foundation via grant IIS-2024775.}}

\maketitle

\begin{abstract}
Robotic systems that can dynamically combine manipulation and locomotion could facilitate dangerous or physically demanding labor. For instance, firefighter humanoid robots could leverage their body by leaning against collapsed building rubble to push it aside. 
Here we introduce a teleoperation system that targets the realization of these tasks using human's whole-body motor skills. We describe a new wheeled humanoid platform, SATYRR, and a novel hands-free teleoperation architecture using a whole-body Human Machine Interface (HMI). This system enables telelocomotion of the humanoid robot using the operator's body motion, freeing their arms for manipulation tasks. In this study we evaluate the efficacy of the proposed system on hardware, and explore the control of SATYRR using two teleoperation mappings that map the operators body pitch and twist to the robot’s velocity or acceleration. Through experiments and user feedback we showcase our preliminary findings of the pilot-system response. Results suggest that the HMI is capable of effectively telelocomoting SATYRR, that pilot preferences should dictate the appropriate motion mapping and gains, and finally that the pilot can better learn to control the system over time. This study represents a fundamental step towards the realization of combined manipulation and locomotion via teleoperation.
\end{abstract}


\IEEEpeerreviewmaketitle

\section{INTRODUCTION}
\IEEEPARstart{H}{umanoid} robots have the potential to aid workers in physically demanding and dangerous jobs, such as firefighting and disaster relief. However, to adequately aid workers in these tasks, robots must be capable of \textbf{D}ynamic \textbf{M}obile \textbf{M}anipulation \textbf{(DMM)}, which is defined as the ability to combine locomotion and manipulation to amplify the forces applied to the environment. These types of tasks are challenging for state-of-the-art robots because they require whole-body coordination, intermittent contacts, large manipulation forces, dynamic locomotion, and a physically capable machine. 

\begin{figure}[t]
\centering 
    \includegraphics[width = \columnwidth]{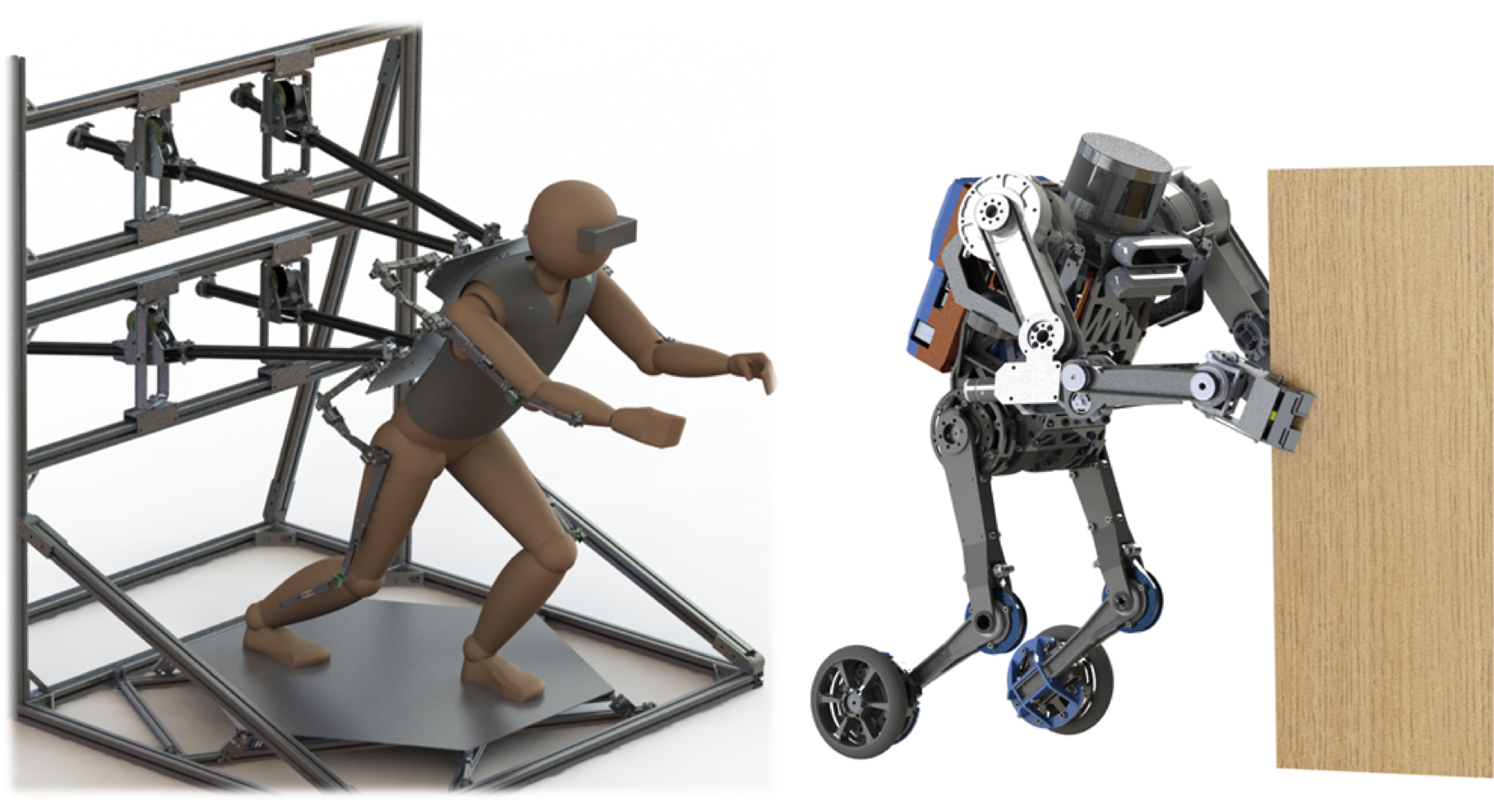}
    \caption{The targeted vision of the teleoperated humanoid platform, SATYRR, performing Dynamic Mobile Manipulation via teleoperation with the Human Machine Interface.}
    \label{ConceptDesign}
\end{figure}
 
Autonomous robots cannot yet match human capability to plan complex actions in unpredictable scenarios inherent to emergency response. To address this, we aim to leverage human sensorimotor skills to directly control the robot's whole-body motion via teleoperation \cite{ProfPaper3,ProfPaper2}, as seen in Fig.\ref{ConceptDesign}. Whole-body humanoid teleoperation can be separated by manipulation and locomotion tasks. In this work, we focus on the study of the latter, telelocomotion, using human whole-body movement.

Existing work on robot locomotion utilizes joysticks, visual-inertial motion camera systems, and human-machine-interfaces. In contrast to these works, \textit{our goal is to create a hands-free strategy for telelocomotion as a fundamental step towards DMM}. We envision that the operator will use their arms to control robot manipulation while intuitively regulating robot locomotion using the pitch and twist of their body. Similar strategies have been used with self-balancing personal transportation devices, such as the Segway \cite{Segway}.

Wheeled bipeds, such as SATYRR, \textit{Ascento} \cite{Ascento1}, and \textit{Handle} \cite{handle}, are the perfect platform to study DMM. Self-balancing wheeled humanoids combine the advantages of legged robots with the agility of wheels. Existing machines have demonstrated nimble navigation over challenging terrain with inclines and curbs, while allowing the exploitation of dynamic upright instability to facilitate DMM tasks. Their human-like design with high center of mass (CoM) and small base of support allows them to exert large manipulation forces to the environment by dynamically throwing their bodies \cite{handle}.

However, dynamic control of wheeled humanoid robots remains a challenge. Due to the highly nonlinear nature of their dynamics, their controls often employ reduced-order models with simpler and more intuitive behavior. For instance, the Linear Inverted Pendulum (LIP) is often used to capture the core dynamics of bipedal locomotion for the planning and control of humanoid robots \cite{LIP}. Similarly, in this work we leverage the \textit{Wheeled Inverted Pendulum (WIP)} model to describe the fundamental behavior of the self-balancing wheeled humanoid SATYRR \cite{Wensing_Paper}. Due to this model's inherent instability, a control framework capable of balancing the robot is required. As seen in \cite{Ascento1, Model_biwheel_decouple}, linear state-space systems and their closed-loop feedback controllers have been effective in stabilizing and enabling locomotion of wheeled bipedal robots.

 The user’s body motion is captured by the HMI (Human-Machine-Interface) \cite{Sunyu_HMI2} and mapped to a desired robot \textit{velocity or acceleration} via a piece-wise continuous mapping and the dynamics of the WIP, as similarly seen in \cite{Sunyu_HMI}. We designed force feedback to prevent the human from falling during teleoperation and enable a larger range of motion. When the operator tilts forward, the translation of the human CoM is mapped into a desired forward motion. When the operator rotates their body, the rotation around the human CoM is mapped into a desired turning rate for SATYRR. 
 In this viability study, we provide three key contributions:
\begin{itemize}
    \item The introduction of a bi-wheeled humanoid robot SATYRR and its integration with a novel whole-body HMI, that enables hands-free telelocomotion.
    \item The development of two human motion mappings (velocity and acceleration) evaluated on two sets of experiments.
    \item Discussions on the viability of the proposed approach and consolidation of user feedback about their human-robot interaction.
\end{itemize}

The experiments conducted help evaluate the efficacy of the proposed system and benchmark the performance. Preliminary results show that our first pilot was only seconds slower when using the HMI compared to using a handheld joystick. The subject also preferred acceleration mapping over velocity mapping. Our second pilot demonstrated their ability to better control the system over three days of trials. These findings suggest that the proposed system is a capable platform for telelocomotion.
 
The remainder of this paper is structured as follows. In section \ref{System Description} we discuss the implementation of hardware for SATYRR and the HMI, along with the key software integration components. In section \ref{ModelControl} we discuss the reduced order model and closed-loop feedback tracking controllers. In section \ref{MotionMappings} the velocity and acceleration based human motion mappings are introduced. Section \ref{experimet} outlines the experiment design. Section \ref{ExperimentDiscussion} discusses the experiment results and user feedback.  Finally, section \ref{concl} provides our conclusion.

\section{System Description} \label{System Description}
\subsection{SATYRR Hardware}
\begin{figure}[t]
    \centering
    \includegraphics[width=\columnwidth]{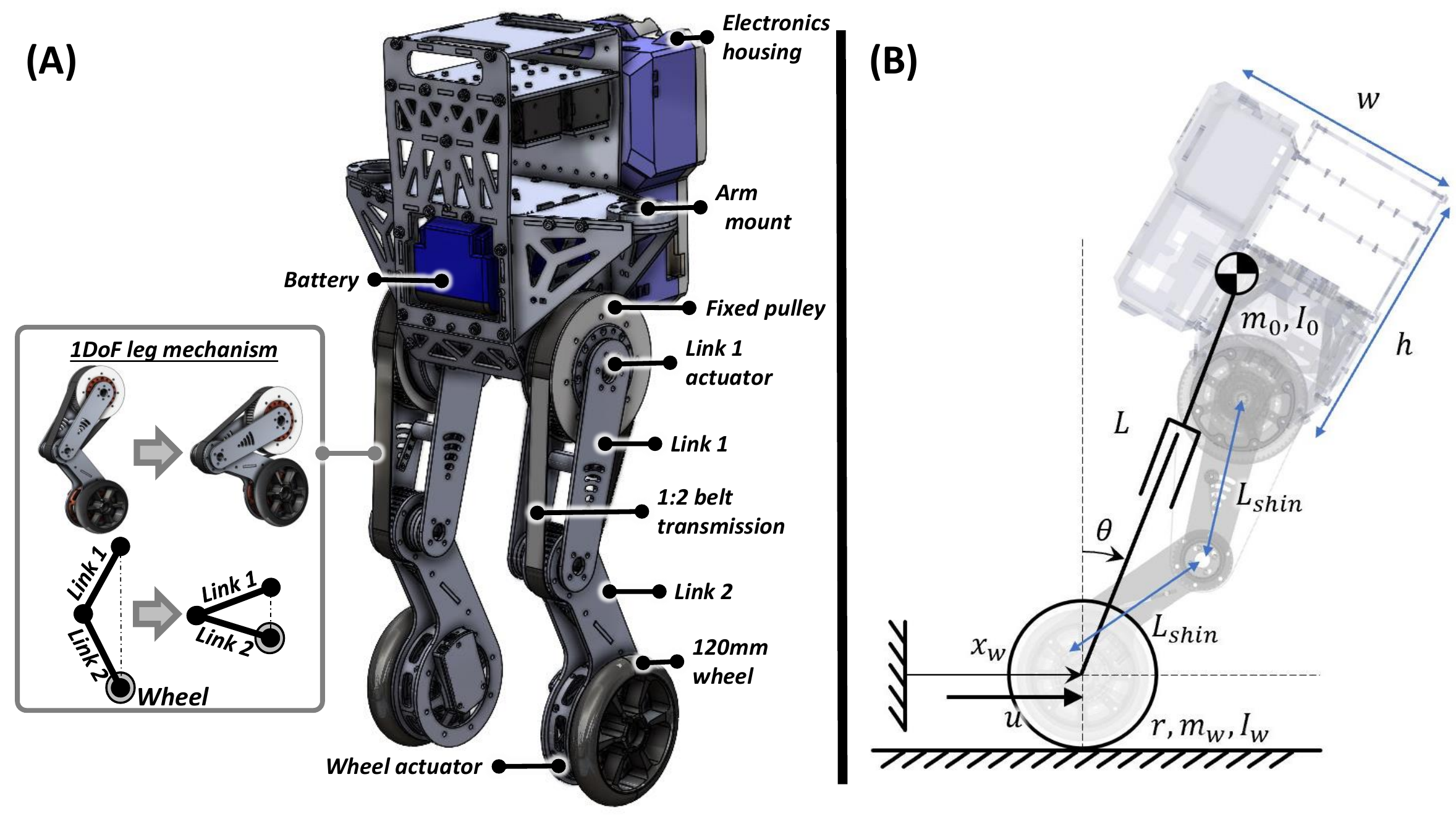}
	\caption{(A) Design of SATYRR's lower body, torso, and straight line mechanism of the single degree-of-freedom leg. (B) WIP reduced order model and states.}
	\label{f_design}
\end{figure}

SATYRR is an anthropomorphic robot with a pair of legs that employ active wheels instead of feet, as shown in Fig. \ref{f_design}. Each leg consists of a one degree-of-freedom (DoF) mechanism that couples links 1 and 2 to move the wheel vertically in a perfect straight line. The actuator directly drives link 1 and a pulley fixed to the torso couples the rotation of link 2, in respect to the robot's body via a 1:2 timing belt transmission. This implies that link 2 will rotate two times faster than link 1, in respect to the robot's torso, resulting in a vertical motion of the wheel because links 1 and 2 have the same length. SATYRR's lower-body is equipped with four of the same actuators utilized in the MIT Mini Cheetah \cite{Katz_thesis,MiniCheetahICRA}. These have a built-in reduction ratio of 6:1 and can deliver a peak torque of 24Nm and achieve speeds up to 30rad/s. The actuators are backdrivable and can indirectly control the output torque via regulation of the current going through the windings \cite{ProprioAct}. The four motors are daisy chained and communicate over CAN bus with the low-level controller sbRIO 9626 from \textit{National Instruments}. Each motor is equipped with a 12-bit encoder and an inertial measurement unit (IMU) VN-100 from \textit{VectorNav} is mounted to the torso. This IMU houses an on-board Kalman filter that outputs the estimation of the robot's torso orientation at a maximum frequency of 200 Hz. To power all the motors, computer, and IMU, a single \textit{Kobalt} 24V and 4Ah battery is used in tandem with two variable low-dropout regulators (LDOs). The system weighs 6.8 kg and stands at a nominal hip height of 0.28 m.
  The robot computer runs the controller in real-time frequency of 833 Hz and communicates with the HMI via UDP.

\subsection{HMI Hardware}
The HMI introduced in \cite{Sunyu_HMI2} and \cite{Sunyu_HMI} is composed of a high force haptic device, a large force plate, and passive Motion Capture (MoCap) exoskeleton.
The haptic device consists of two backdrivable linear actuators that are capable of generating up to 100N of force each to the operator's torso near the CoM. Each actuator is mounted on a a passive gimbal and measures the spatial location of the operator's torso. The actuators are utilized in our experiments to create a spring-like force that allows the operator to tilt their body forward and backward without falling. Future versions of this device will employ four linear actuators (Fig. \ref{ConceptDesign}) and allow the generation of forces and torques to the operator's torso in multiple directions. 
The design of the force plate is similar to that of a \textit{Stewart Platform} \cite{Stewart} but with six uniaxial load cells, instead of linear actuators. It measures the magnitude and location of the net ground contact force applied by the operator's feet. Its point of application is known as the Center of Pressure (CoP). 

The HMI is controlled by a real-time controller (cRIO-9082 from \textit{National Instruments}) that runs the at 1kHz.
The HMI can be used for providing feedback about disturbances to the robot’s dynamics to the user, which will significantly improve teleoperation performance in the future. For example, if a disturbance force is applied to the robot, the HMI applies a similar force to the user, who uses their balancing reflexes for stabilizing locomotion. User's dynamics measured by the force plate and exoskeleton system can be used to calculate the feedback force for the user comparing it to the robot's dynamics. This strategy has been employed in the author's previous work on whole-body teleoperation of humanoid robots \cite{ProfPaper1,Ramos_SciRob}.

\begin{figure*}[t] 
    \centerline{\includegraphics[width=18.5cm]{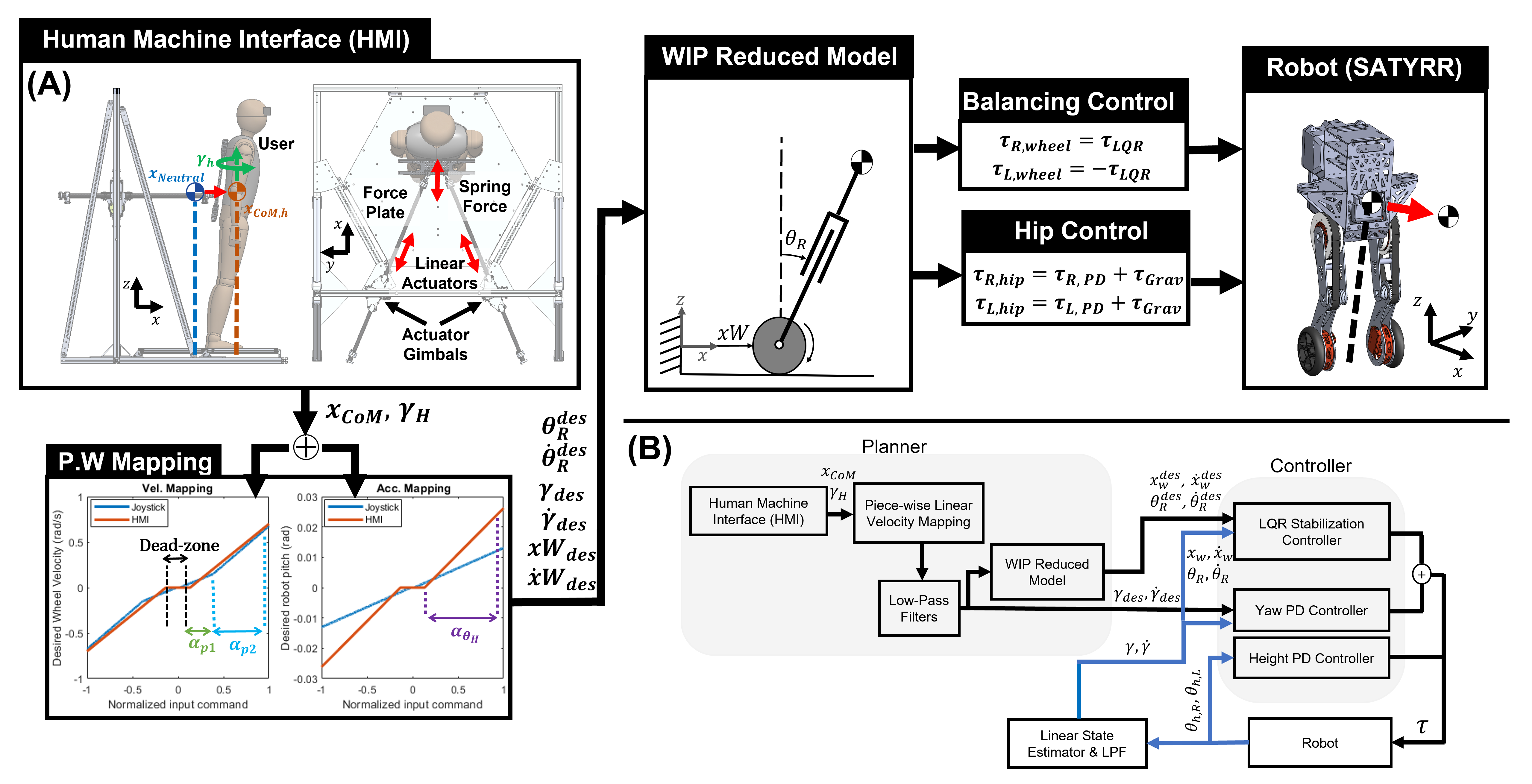}}
    \caption{(A) Full system layout and control flow work from pilot to the WIP to the full robot. (B) Overview of the control architecture.}
    \label{figHMI}
\end{figure*}

In this study, we introduce a simple unilateral control strategy for the preliminary investigation of the human-robot control system. The position of the user's CoM is measured by the linear actuators and utilized for the control of robot motion in its sagittal plane. The turn rate of the robot is controlled by the twist angle of the human along the longitudinal $z$-axis as shown in Fig. \ref{figHMI}, and described in Section \ref{ModelControl}.


\section{Modeling \& Controller Design} 
\label{ModelControl}
In section \ref{Modeling} we discuss the planar WIP model that reduces the dimensionality of the system while still enabling driving and steering. In section \ref{StabilizationController} we discuss the feedback controller that handles the dynamic balance of the robot around the desired upright equilibrium. Section \ref{HeightController} discusses additional controllers that regulate the robot height (leg length) and turning rate (yaw) around the vertical $Z$ axis.

\subsection{The Wheeled Inverted Pendulum (WIP) Model}
\label{Modeling}
As seen in \cite{Ascento1}, \cite{Wensing_Paper}, and \cite{CartPole_Model}, reduced order models (RoM) of self-balancing wheeled systems have been effective for motion planning and dynamic stabilization. We use the planar WIP model, composed of active wheels and the lumped rigid-body that represents the robot's torso, as shown in Fig. \ref{figHMI}(B), to develop the stabilization controller. 
Moreover, we assume no slip between the wheels and the ground, which implies that the distance traveled, $x_w$, is linearly proportional to the total number of wheel revolutions. 
The dynamic equations of motion of the WIP are given by:
\begin{equation}\label{wip_dyn}
    \begin{gathered}[b] 
        \bigg (m_{o} \!+\! m_{w} \!+\! \frac{I_w}{r^2}\bigg )\ddot{x}_w \!+\! m_{o}L s_{\theta_R}\dot{\theta}^2_R 
    \!-\! m_{o}L c_{\theta_R}\ddot{\theta}_R \!= \!u\\
        (m_{o}L^2 + I_{o})\ddot{\theta}_R - m_{o}Lc_{\theta_R}\ddot{x}_w - m_{o} gLs_{\theta_R} \!= \!0, 
    \end{gathered}\
\end{equation}
where $m_{b}$ is the mass of the body, $m_w$ is the wheel mass, $L$ is the length between the center of the wheel and the CoM of the body, $\theta_b$ is the pitch rotation of the body from vertical axis, $I_w$ is the inertia of the wheel, $I_{b}$ is the inertia of the body evaluated at its CoM, $r$ is the radius of the wheel, and $u$ is the control effort (force) exerted along the $x-$axis on the wheel due to the motor torque. 

For the controller synthesis, we acquire the linear state-space model by linearizing equation \eqref{wip_dyn} around the upright equilibrium state $q_0$ at the nominal center of mass height.
\begin{align*}
    \delta \dot{q} = A\delta q + B\delta u.
\end{align*}
Where $q = [x_w \quad \theta_R \quad \dot{x}_w \quad \dot{\theta}_R]^\top$ is the state vector; $A \in \R^{n\times n}$, $B \in \R^{n \times k}$ are the state-space matrices, $\delta q = q-q_0 \in \R^n$ is the deviation from the equilibrium state $q_0$ and $\delta u=u-u_0\in \R^k$ is the deviation from the nominal control effort $u_0$. Where $n = 4$ and $k = 1$. 

\subsection{Stabilization Controller}
\label{StabilizationController}
The stabilization task can be simplified by leveraging the mechanical design of our robot, which constrains the CoM to remain vertically aligned with the center of the wheel. This feature enables decoupling of the posture control to regulate height of the robot, the stabilization to keep SATYRR balanced, and the steering controller \cite{Wensing_Paper}, \cite{Model_biwheel_decouple}.

By discretizing the continuous state-space model with a sampling period $T_s = 1.0$ ms and assuming an infinite horizon, we can solve the discrete time Algebraic Riccati Equation for the corresponding optimal gain matrix $K$ \cite{franklin1998digital}. After extensive tuning of the cost matrices, the optimal gains generated by the model were $K= [-180, -640, -120, -70]$. 


To control the real robot using our planar model, we assumed that the legs behave symmetrically. The left leg, right leg hip, and wheel angles are averaged to determine robot height and wheel displacement. Final stage mapping can be seen in Fig. \ref{figHMI}. 


\subsection{Height Adjustment \& Yaw Controller}
\label{HeightController}
The height adjustment PD controller is given by:
\begin{equation}
    \tau_{Hip} = J_c^\top F_g + K_p(\theta_{1, des} - \theta_1) + K_d(\dot{\theta}_{1, des} - \dot{\theta}_1),
\end{equation}
where $\theta_{1}$ represents the angle between the torso and link 1, $F_g=m_0g$ represents the force due to gravity, and $J_c$ is the Jacobian that relates the relative angular velocity $\dot{\theta}_1$ of link 1 with the vertical velocity of the foot. Tuned on hardware, $K_p = 100$ and $K_d = 1$. 

The robot's heading PD controller is given by:
\begin{equation}
    \tau_{yaw} = K_{p,yaw}(\gamma_{des} - \gamma) + K_{d,yaw}(\dot{\gamma}_{des} - \dot{\gamma}). 
\end{equation}
Where $\gamma$ represents the robot's yaw, and $\theta_{w,R}, \theta_{w,L}$ represent the left and right wheel encoder positions:
\begin{equation}
    \gamma = \frac{r_w}{r_c}(\theta_{w,R} - \theta_{w,L}).
\end{equation}
 Here $r_w$ represents the wheel radius and $r_c$ represents the distance between the two wheels. Hand tuned on hardware, $K_{p,yaw} = 1$ and $K_{d,yaw} = 0.1$.

\section{Locomotion Teleoperation Mappings} \label{MotionMappings}
 For intuitive hands-free locomotion control of SATYRR, we explore the efficacy of two, hand tuned piece-wise linear mappings that map tilt to the robot WIP's velocity or acceleration.The two mappings differ in the type of control authority they give the user \cite{Sunyu_HMI}. Section \ref{HumanMotionVelocityMapping} discusses a proposed velocity mapping. Section \ref{HumanMotionAccMapping} outlines how the user's tilt is mapped to the robots tilt and acceleration.

\subsection{Human Motion Velocity Mapping}
\label{HumanMotionVelocityMapping}
 To control linear velocity of the robot, a joystick's tilt or user’s CoM displacement along the $x$ axis, $p_x$, is mapped to a desired translation velocity of the robot's wheel, $\dot{x}^{des}_{w}$. We employ a piece-wise linear transformation with a small dead-band, $p^{db}$. This dead-band accounts for natural human swaying or joystick noise while at home position. The mappings, as seen in Fig. \ref{figHMI}, enable the pilot to fine tune their desired commands and robot response. The $x$-directional virtual spring rendered by the HMI's linear actuator allows the operator to lean further forward or backwards without falling. The spring constant was set by user's preference for desired stiffness and maximum pitch. 
 
The piece-wise linear mapping function is defined as:
\begin{equation} \label{piecewise_map}
    \dot{x}^{des}_{w} \!\!=\!\!
    \begin{cases} 
      0 & |p_x| < p^{db}_x \\
      sgn(p_x)\alpha_{p_1}(|p_x|\!-\!p^{db}_{x}) & p^{db}_{x}\!\! \leq \!\! |p_x| \!\! <\!\! p^{swp}_{x} \\
      sgn(p_x)\alpha_{p_2}(|p_x|\!-\!p^{swp}_{x}) \!+ \!c^{swp} & p^{swp}_{x}\!\! \leq \!\! |p_x| \!\!<\!\! p^{max}_{x} \\
      sgn(p_x)\dot{x}^{max}_{w} & otherwise
  \end{cases}
\end{equation}
where $\alpha_{p_1}, \alpha_{p_2}$ are the slopes of the individual piece-wise linear sections. $p^{swp}$ represents the switching coordinate, and $c^{swp}$ represents the intersection of the two piece-wise sections that ensures continuity. 
The $sgn()$ function returns a binary value $[-1 \quad 1]$ depending on the sign of the input.

The turning rate of the robot is analogously controlled by the rotational angle of the human along $z$-axis, $\gamma_{h}$. The angle is mapped to a desired yaw rate, $\dot{\gamma}_{des}$, using a similar piece-wise linear function but with different coefficients. 
By controlling driving velocity and turning rate using their whole-body, the operator's hands are left free for future telemanipulation tasks.

\subsection{Human Motion Acceleration Mapping}
 \label{HumanMotionAccMapping}
 The acceleration mapping (\ref{piecewise_acc_map}) maps the joystick tilt or human tilt, $\theta_H$, to the robot's desired WIP tilt, $\theta_R^{des}$. Since the WIP's tilt is proportional to its CoM acceleration, this mapping can be understood as commanding acceleration. Under the small-angle assumption, the human's CoM displacement along the y axis is approximated as their tilt, $p_y \approx \theta_H$. Accordingly, the piece-wise mapping function is given by:
\begin{equation}\label{piecewise_acc_map}
    \theta_R^{des} \!\!=\!\!
    \begin{cases} 
      0 & |\theta_H| < \theta_H^{db} \\
      sgn(\theta_H)\alpha_{\theta_H}(|\theta_H|\!-\!\theta_H^{db}) & \theta^{db}_{H}\!\! \leq \!\! |\theta_H| \!\! <\!\! \theta^{max}_{H} \\
      sgn(\theta_H)\theta^{max}_{R} & else
  \end{cases}
\end{equation}
where $\theta_H^{db}$ represents a dead-band around zero, $\alpha_{\theta_H}$ is the section slope, and $\theta_{R}^{max}$ represents the maximum commanded tilt.
Differentiating this input we find $\ddot{\theta}_R^{des}$. Under the small angle approximation  ($sin(\theta_R)\approx\theta_R ;\, \cos(\theta)\approx 1$), and assuming small pendulum velocities ($\dot{\theta_R^2} \approx 0$), we rearrange (\ref{wip_dyn}) to: 

\begin{equation}\label{accEqn}
     \ddot{x}_w^{des} = -\frac{(m_{0}L^2 + I_{0})}{m_{0}L}\ddot{\theta_R^{des}} + g\theta_R^{des}
\end{equation}

Integrating (\ref{accEqn}) we find the the desired states, $\dot{x}_w^{des}$ and $x_w^{des}$, for our stabilization controller. 
\section{Experimental Design}
\label{experimet}
\begin{figure}[h] 
    \centering
	\includegraphics[width=\columnwidth]{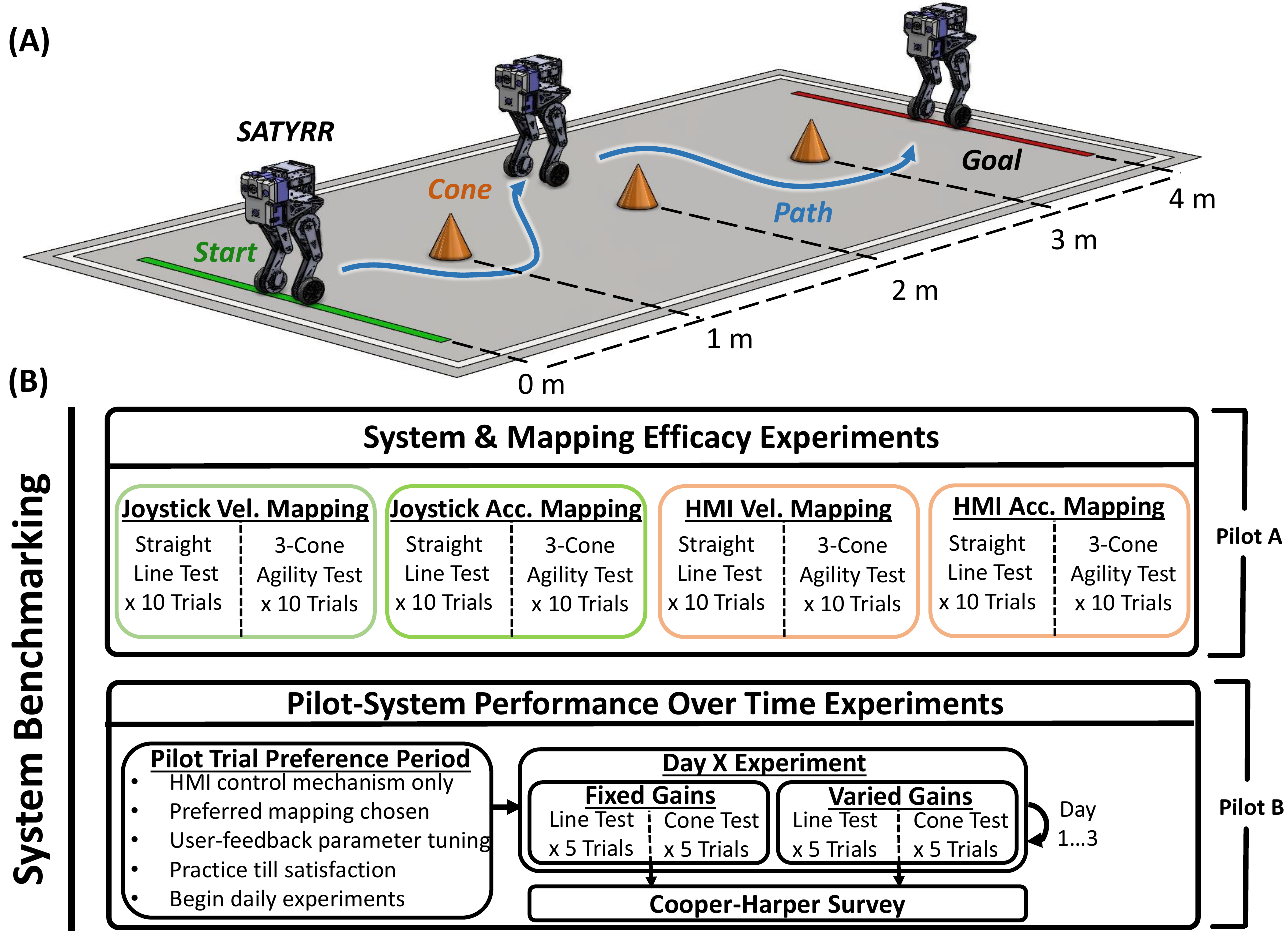}
	\caption{(A) Obstacle avoidance experiment test course. The linear motion test was setup in the same area without the cones. (B) The two experimental designs and layout for benchmarking.}
	\label{f_expLayout}
\end{figure}

The goal of the experiments is to demonstrate the integration of SATYRR with the HMI, and perform benchmark evaluations to determine the viability of different human-robot-motion mappings for telelocomotion. Two sets of human subject experiments were developed: 1) System and mapping efficacy experiments and 2) Pilot-system response experiments. Inspired by sports-based agility tests \cite{DevelopingAgilityQuickness,stewart2014reliability}, two test courses were designed: A) Straight-line speed course and B) 3-cone drill course as seen in Fig. \ref{f_expLayout}. These tests focus on demonstrating rapid linear motion and confined navigation around obstacles, respectfully. Each experiment (1 or 2) required the completion of these two courses (A+B). The experiment layout can be seen in Fig. \ref{f_expLayout}.


The goal of both tests is to reach the end destination in the shortest time possible. The second test introduced obstacles (cones) that the user is also required to avoid colliding with by going through the cones in a weave-like pattern. The successful completion of a single run is defined as the translation from start to goal line following the design of the test. Instability caused by collisions while steering, venturing outside the predefined region, and slipping/falling due to user error are qualified as failed runs. Moreover, the user is required to retain stable control of the robot upon passing of the goal line. 
Experiment 1 required Pilot A to test all eight combinations of tests (e.g., joystick-velocity-mapping-cone test is a single combination), as can be seen in in Fig. \ref{f_expLayout}. The user was allowed to practice with each combination until satisfaction but only allowed to change their mapping and controller gains for different controller-mechanism-mapping schemes (e.g., the gains using the joystick for velocity mapping were fixed for its straight-line and cone drill test). This enforces that the user choose gains capable of telelocomoting the robot along open straight paths and around obstacles since a priori knowledge of the terrain would not be known in practicality. After tuning and practicing, the pilot was required to complete 10 successful completion runs for each of the 8 combinations. Each combination took between 3-3.5 hours to tune, practice, and complete.

Experiment 2 required Pilot B to choose a preferred motion mapping for their tests. After an initial trial preference period, pilot B chose velocity mapping. Their mapping gains were adjusted prior to day 1 and set as their fixed gain set – unmodifiable over the next 3 days. The variable gain set could be adjusted and tested daily prior to starting. Over the 3 days, Pilot B was required to complete 5 successful runs with their fixed gains and variable gains for both the straight-line test and 3-cone agility test.

For these experiments the robot CoM height is fixed at 0.28m. The pilot is given a first-person view of the robot during the experiments through DJI FPV googles and a single monocular camera on the robot. The camera orientation is fixed to align with the center of the robot and pitched as per the user’s preference prior to the start of the tests. For both cone and straight-line tests, the robot is started from the same position 4m away from the finish line. The human’s task is to teleoperate the robot after a 3 second countdown. 

Due to the lengthy duration of the experiments and because this work represents a viability study only two subjects of age 32 years – pilot A – and 21 years - pilot B – were recruited. Neither pilot had competitive athletic involvement beyond high school.
These experiments were conducted after the University's Institutional Review Board reviewed and approved this research study.

\section{Experimental Results and Discussion}
Sections A, and B, discuss the results of the system efficacy experiments with pilot A. Section C discusses the key points from the second set of experiments with pilot B. Section D consolidates feedback from both pilots. 
\label{ExperimentDiscussion}

\subsection{HMI Control Efficacy}
\begin{figure*}[ht]
    \centerline{\includegraphics[width=18.5cm]{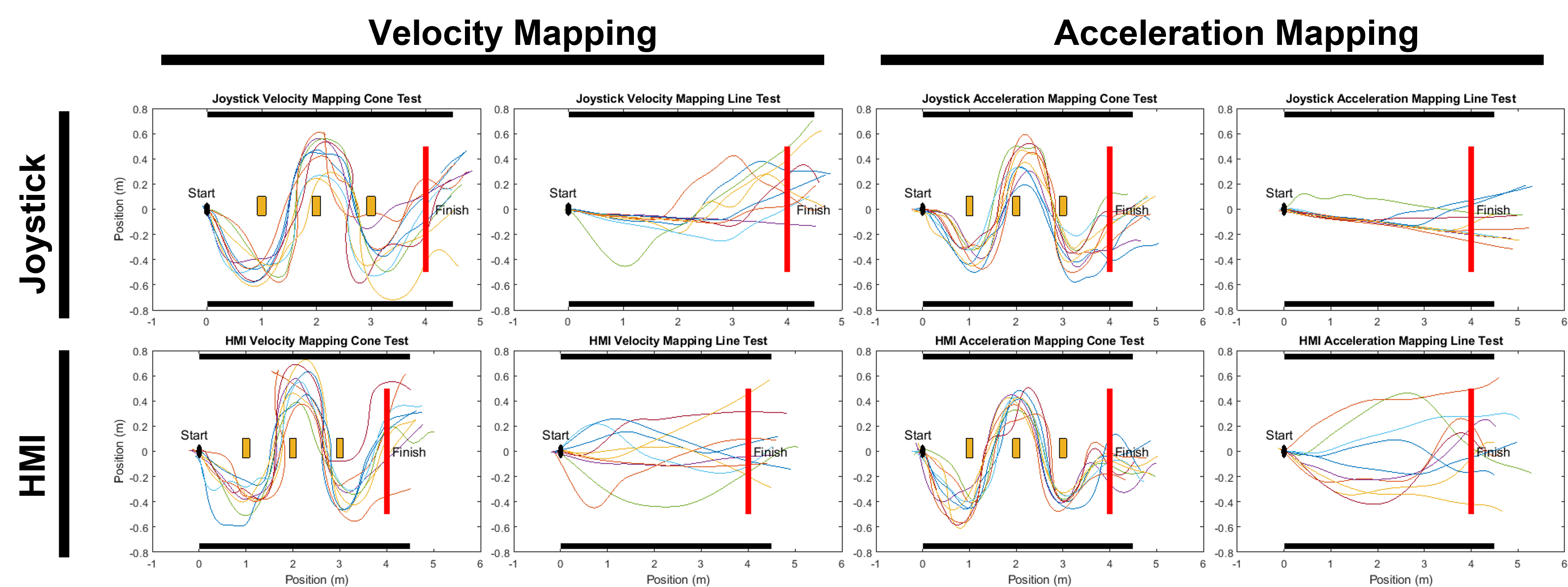}}
    \caption{Reconstructed odometry data for all 8 combination test runs for the system efficacy experiments with pilot A. Each combination shows the first 10 successful runs.}
    \label{f_odometry}
\end{figure*}

To test the practicality of using the HMI with feedback force for hands-free tele-locomotion, we use the conventional joystick as a gold-standard for comparison. Across the same mappings we notice that pilot A's performance was superior using the joystick. The fastest time across the obstacle test was completed using the joystick at 9.2 seconds. However, the user had the fastest performance along the straight-line test using the HMI in 5.9 seconds. The average total distance traveled along the joystick-cone test was 5.03m and 5.18m for the HMI-cone test – a difference of 150 cm. Fig. \ref{f_odometry} highlights the paths taken by the robot using the joystick and HMI for all 10 successful run. Note that the robot's path was reconstructed using wheel odometry, which is prone to error accumulation due to wheel slip.

We hypothesize three reasons the user may have been slower with the HMI: 1) There may exist a difference in experience between operating the two systems as most people are likely more familiar with the usage of a handheld joystick. 2) The pilot noted that during the HMI experiments, first-person view dizziness was augmented by mismatch between their body motion and the robot’s direction of motion. To mitigate dizziness the pilot may have unintentionally modulated their control with more conservative body movements and velocities. This pilot also felt less dizzy when standing stationary using the joystick. 3) Finger motion has an inherently faster bandwidth in comparison to the movement of the operator’s whole-body. This results in a slower rise time of the desired velocities commanded by the pilot and subsequently a slower response from the robot.



These preliminary results indicate that the HMI is only seconds slower than the completion time when using the joystick, and the difference in average path lengths differs by centimeters. This supports the assumption that the operator can use their body to control the robot with almost the same dexterity as with their hands.

\begin{figure}[ht] 
    \centering
	\includegraphics[width=\columnwidth]{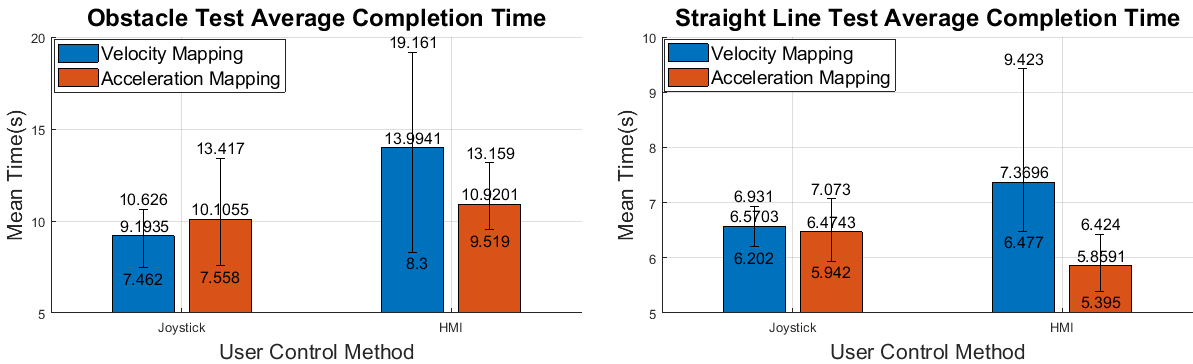}
	\caption{Experiment 1 average completion time for pilot A on 10 successful test runs. Error bars show max and min values for each of 8 combinations.}
	\label{f_AvgTimePlot}
\end{figure}

\begin{figure*}[ht]
    \centering
	\includegraphics[width=17.5cm]{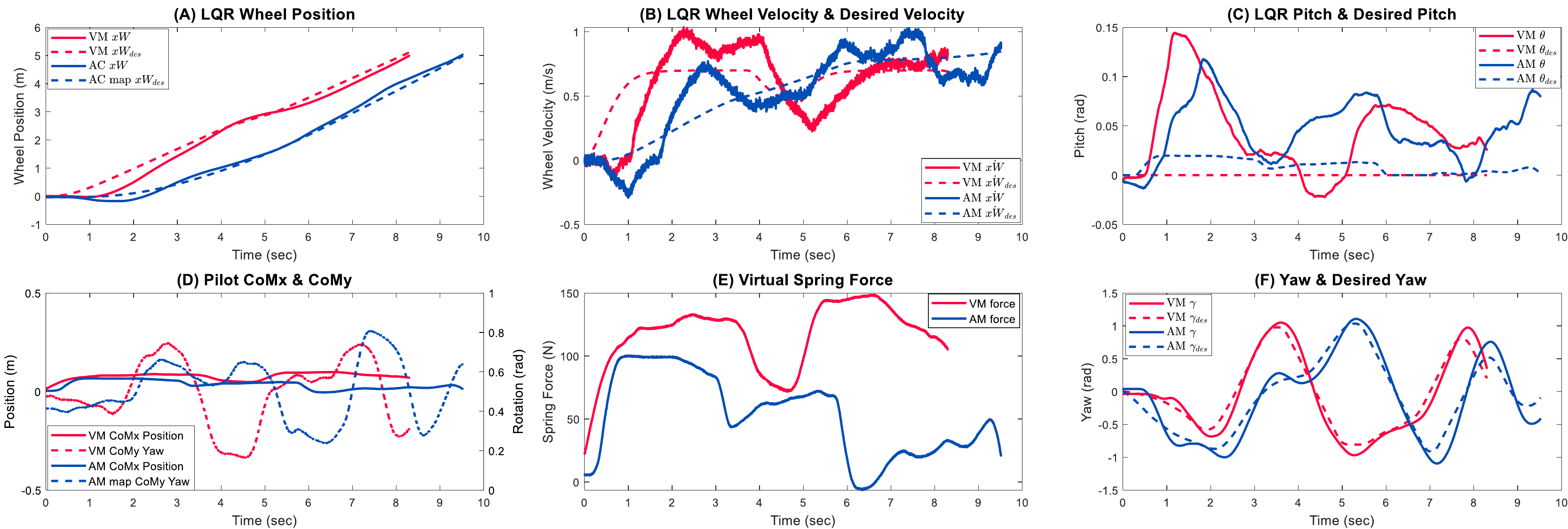}
	\caption{HMI single-run 3-Cone obstacle avoidance agility test: actual and desired states, applied torques, and pilot CoM position for velocity and acceleration mapping. (A) and (B) show characteristic behavior of non-minmum phase systems e.g pendulums - the robot moves backward initially before moving forward. Legend guide: VM - velocity mapping. AM - acceleration mapping}
	\label{f_allData_lin}
\end{figure*}
  
\subsection{Human Motion Mapping Preference}
Pilot A performed the experiments more quickly using the acceleration mapping as 3 of the 4 test setups resulted in lower time, as seen in Fig. \ref{f_AvgTimePlot}.  The fastest completion time recorded for any test was using the HMI-acceleration mapping at $5.9$ seconds. This strategy enables the user to move faster than they could have with the velocity mapping, but in turn places responsibility on them to stabilize the robot – continuous acceleration causes the robot to pitch too far and fall.  We hypothesize 2 reasons that the velocity mapping may have resulted in slower times: 1) The user's commanded inputs were perturbed by small oscillations of the robot making velocity control more difficult since directly commanding yawing and forward motion are coupled dynamics.  These oscillations ($\approx\pm 0.075$m in amplitude) around the desired wheel position occur at slower speeds and while the robot is trying to stay stationary. They are caused by a combination of actuation delay and friction in the wheels. 2) Pilot A's natural control preference leans towards control of acceleration rather than velocity. 

Our observations suggest a strategy for constructing these piece-wise mapping functions. For both mappings, the subject desired slower and more precise movements at lower angles of tilt for the obstacle avoidance test but more aggressive gains – i.e. higher achievable velocities or tilt – for the straight line test where the user throttled maximum input. For the velocity mapping this dictates a moderate gain within the first section of the piece-wise function and higher gains for the second section. Note, creating a function with widely different gains can result in jerkiness in the robot’s response and thus requires careful tuning with user feedback. For the acceleration mapping, a simple linear function with small gains can be used – pilot A preferred mapping their maximum pitch to a desired max $1.5^o$ tilt of the robot. We believe there are 2 reasons for this small gain preference: 1) The robot’s acceleration is a function of the pilot’s pitch and pitch acceleration shown in \eqref{accEqn}. This acceleration is significant as the user is never truly standing still in the HMI. 2) Feedback from the user also revealed the key to their acceleration mapping control strategy – accelerating to a desired constant velocity, then commanding zero acceleration, as seen by the virual spring force in Fig.\ref{f_allData_lin}E, and using quick yawing motions to correct for the robot’s position. Smaller acceleration gains would result in a slower velocity increase and prevent the user from overshooting their desired robot speed. Note that with this mapping, the robot velocity is not limited for the straight-line test. 

The velocity mapping had 7 failed runs with the joystick, and 11 failed runs with the HMI before the completion of the 10 successful runs – a total of 18 failed runs. The acceleration mapping had 3 failed runs with the joystick and 4 failed runs with the HMI – a total of 7 failed runs. The completion times and consistency of the acceleration mapping serve as early indicators that our pilot preferred acceleration mapping.  The testing conducted on hardware indicates small timing differences – seconds apart – between the different mappings. This supports our claim that both mappings present viable approaches to control of SATYYR, and pilots may indeed have preferences in control mapping schemes. 

\subsection{Does the Pilot improve with time?}
Implicitly assumed in our goal towards whole-body teleoperation is the idea that the pilot can learn to operate the robot more effectively through adaptation in their control. Through our preliminary findings we show here that this assumption is well founded and that pilots can improve their course telelocomotion performance by practicing and altering their body motion. 

\begin{figure}[ht] 
    \centering
	\includegraphics[width=\columnwidth]{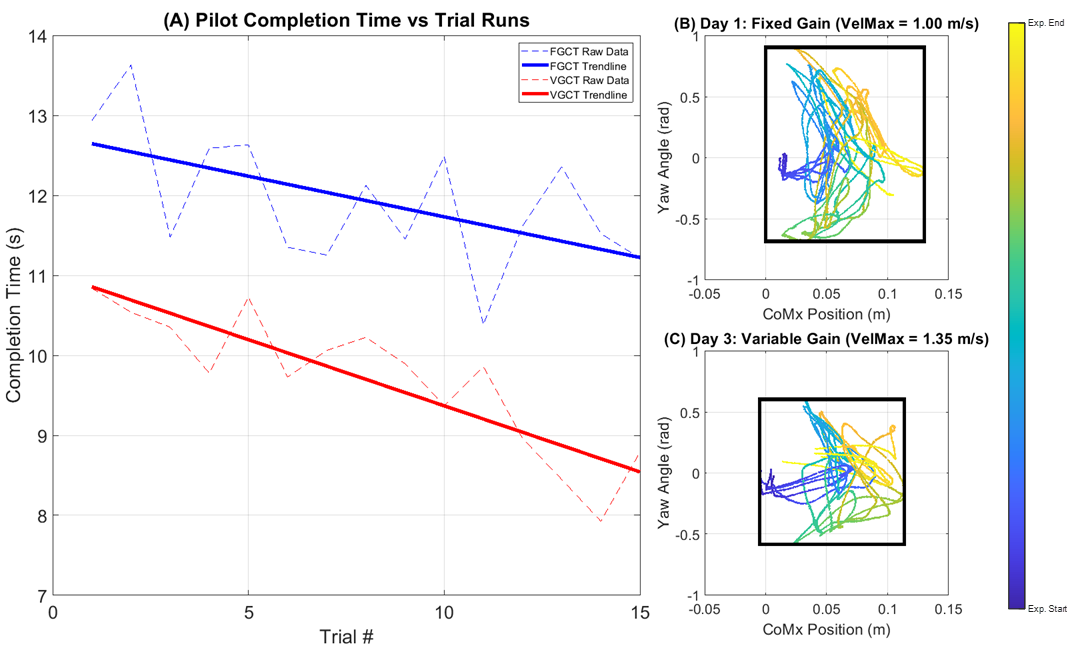}
	\caption{(A) Performance over time data. Fixed-Gain-Cone-Test (FGCT). Variable-Gain-Cone-Test (VGCT). (B) User human motion area for Day 1. (C) User human motion area for Day 3.}
	\label{f_TimePerf}
\end{figure}

Fig. \ref{f_TimePerf}(A) illustrates comparisons between two sets of gains – fixed gains and varied gains – over time. The user-preferred fixed gain maximum velocity was 1.0 m/s. Over the duration of the experiments, the varied mappings gains were increased by 10\%, 25\% and 35\% by user preference. The average completion time of D1-FGCT (Day 1 Fixed-Gain-Cone-Test) was 12.65 seconds, and 11.41 seconds for D3-FGCT – a decrease of 9.8\% over 15 trials. The average completion time of D3-VGCT was 8.79 seconds. There was no significant difference in the completion times of the straight-line tests. From these results we see that the user was able to complete the 3-cone agility test more quickly over the 3 days and consistently showed better performance with the higher varied gains. We believe that for the straight-line test, which requires less movement and skill, the pilot was limited by the achievable maximum velocity of the robot – 1.4 m/s.

For velocity mapping (chosen by the pilot) we define human- control sensitivity, $S$ as:
\begin{equation*}
    S = \frac{\dot{x}^{max}_{w}}{\theta^{max}_{H}}
\end{equation*}

The maximum pitch, $\theta^{max}_{H}$, is determined by the user’s comfort prior to the start of the experiments and is fixed. The preferred maximum velocity is user set and seen in Eqn. (\ref{piecewise_map}). In the pilot’s attempt to complete the course faster, they desired a greater maximum velocity. This leads to an increase in the sensitivity of the system. This pilot adjusted for this increased sensitivity by using smaller, more acute body motions as seen by the reduction in the used human motion area in Fig. \ref{f_TimePerf}(B)-(C). We noticed this emerging behavior with other pilots who operated the system as well.  This may suggest that with long-term training, velocity mapping users may converge to a high-gain-small-tilt strategy \cite{Sunyu_HMI}. To address our initial question and benchmark the systems development over time, we conclude that the pilot can adapt and learn to better control this system. Future studies look to better understand the limitations of this human-robot learning and generalize these preliminary result for larger sample sizes.

\subsection{Human-Robot Interaction}

Both pilots reported that the usage of the joystick and HMI for teleoperation resulted in some disorientation. As discussed above, we believe this problem was a byproduct of human-robot motion mismatch while using the HMI and robot pendulum-like swaying motion. 
Also, without a 3rd person view of the robot, both pilots expressed concern about knowing where the wheels of the robot relative to their view and were sometimes unaware they had hit the cones. This exposes a potential need for a more interactive and user-friendly interface to perform more dynamic tests in the future. 

With larger LQR tracking gains, the swaying motions were exasperated as the robot was more aggressive in its response to the pilot’s command. Sudden stops and turns resulted in some overshoot, that led to a quicker and more aggressive pendulum like motion response from the robot. As such, both pilots requested that the wheel position tracking gains of the stabilization controller be reduced significantly. These findings suggest that our stabilization controller with a human in the loop may benefit from preference based control designs \cite{tucker2021preference}.

These adjusted controller gains, user field-of-view, and swaying dizziness may have led to sub-optimal robot paths, seen in Fig. \ref{f_odometry}. While increased yawing sensitivity was crucial for the obstacle avoidance test, the linear motion test would require less aggressive gains for yawing, suggesting that this mapping should depend on the expected path and desired forward velocity. Finally, both subjects did report feeling more comfortable with the viewpoint and control of the robot with increased practice and test runs. This provides a positive indicator that further training of the operators with the HMI could result in increased performance.

\subsection{Limitations of the Study \& Future Work}
The experiments conducted suggest the \textit{viability} of the hands-free approach for telelocomotion and were not designed to search for optimal performance. Different motion mappings could lead to improved results. In addition, due to the complexity of the system and the extensive testing required from the pilots, the results for only two operators are shown in this work. Moreover, due to physical limitations of the robot – its wheel size and maximum motor speed – we could not achieve speeds beyond 1.4 m/s. Moving forward we look to employ head tracking to reduce user motion dizziness and a pair of arms for manipulation. We believe the added payload can be compensated by the balancing controller to minimally degrade human performance, \cite{handle}. Finally, we look forward to integrating bilateral feedback \cite{ProfPaper3} into our design as to better achieve dynamic similarity between the user and robot.   
\section{Conclusions}
\label{concl}

The contributions presented here are: 1) Introduction of a bi-wheeled humanoid, SATYRR, and its integration with the HMI towards whole-body teleoperation.  2) Completion of benchmark experiments to test robot agility and telelocomotion by using body pitch. For contribution 1) we show the hardware setup, reduced order model, tracking controllers, and motion mappings that enabled hands-free teleoperation of SATYRR. For contribution 2) we ran 2 experiments. In experiment 1 we used completion time (CT), average path length traveled, and number of failed runs to address the efficacy of the proposed system with different mapping types, and subject preferences. In experiment 2 we fix the mapping and look at used motion area and CT over 3 days to determine if the pilot can adjust and learn to operate better. The results suggest that the HMI is a capable platform for controlling SATYRR, that motion mapping preferences are dictated by the pilot, and that users can show improved performance by practicing and adapting their motion behaviors over time. Finally, we discuss feedback from the pilots to better understand the limitations of the system. These preliminary results suggest viability and support future developments of a hands-free telelocomotion approach.


\appendices
\bibliographystyle{IEEEtran}
\bibliography{Main.bib}
\end{document}